\documentclass[conference]{ieeeconf}

\IEEEoverridecommandlockouts
\usepackage{amsmath,amssymb,amsfonts}
\usepackage{algorithmic}
\usepackage{graphicx}
\usepackage{textcomp}
\usepackage{xcolor}
\usepackage{booktabs}
\usepackage{multirow}
\usepackage{verbatim}
\usepackage{afterpage} 
\makeatletter
\let\NAT@parse\undefined
\makeatother
\usepackage{hyperref}
\hypersetup{hidelinks}
\usepackage{lipsum} 
\usepackage{tabularx} 
\usepackage{ragged2e} 
\usepackage[export]{adjustbox}
\usepackage{booktabs}
\usepackage{pifont}
\newcommand{\cmark}{\ding{51}}
\newcommand{\xmark}{\ding{55}}

\usepackage{caption}
\renewcommand{\thesection}{\arabic{section}}
\renewcommand{\thesubsection}{\thesection.\arabic{subsection}}
\renewcommand{\thesectiondis}{\thesection.}

\renewcommand{\thetable}{\arabic{table}}

\newcommand{\bk}{\discretionary{}{}{}}

\usepackage{color}

\newcommand{\ssection}[1]{\noindent {\bf #1} }

\usepackage{listings}
\definecolor{codegreen}{rgb}{0,0.6,0}
\definecolor{codegray}{rgb}{0.5,0.5,0.5}
\definecolor{codepurple}{rgb}{0.58,0,0.82}
\definecolor{backcolour}{rgb}{0.95,0.95,0.92}

\lstdefinestyle{mystyle}{
  backgroundcolor=\color{backcolour}, commentstyle=\color{codegreen},
  keywordstyle=\color{magenta},
  numberstyle=\tiny\color{codegray},
  stringstyle=\color{codepurple},
  basicstyle=\ttfamily\footnotesize,
  breakatwhitespace=false,         
  breaklines=true,                 
  captionpos=b,                    
  keepspaces=true,                 
  numbers=left,                    
  numbersep=5pt,                  
  showspaces=false,                
  showstringspaces=false,
  showtabs=false,                  
  tabsize=2
}

\def\BibTeX{{\rm B\kern-.05em{\sc i\kern-.025em b}\kern-.08em
    T\kern-.1667em\lower.7ex\hbox{E}\kern-.125emX}}

\newcolumntype{L}{>{\RaggedRight\arraybackslash}X}

\begin{document}

\title{\LARGE \bf EgoKit: Towards Unified Low-Cost Egocentric Data Collection with Heterogeneous Devices}

\author{%
\begin{tabular}{c}
Liuchuan Yu$^{1\dagger}$, Erdem Murat$^{1}$, Beichen Wang$^{1}$, Yan Zeng$^{2}$, Tingting Luo$^{2}$, \\ 
Huizhen Zhou$^{1}$, Shanghao Li$^{3}$, Huining Feng$^{1}$, Zhigen Zhao$^{4}$,\\
Ning Yang$^{5}$, Ke Jing$^{\ddagger}$, Yunhao Liu$^{6}$, Ruoya Sheng$^{\S}$
\end{tabular}
\thanks{$^{\dagger}$George Mason University, Computer Science, Fairfax, Virginia 22030, USA. Send correspondence to {\tt\small $^{\dagger}$liuchuany@acm.org, $^{\ddagger}$ke.jing@bytedance.com, $^{\S}$ruoya.sheng@gmail.com}}%
\thanks{$^{1}${\tt\small \{lyu20,emurat,bwang25,hzhou9,hfeng2\}@gmu.edu}}
\thanks{$^{2}${\tt\small \{tingtingluo.ux,jenniferzeng5562\}@gmail.com}}
\thanks{$^{3}${\tt\small sli261@uic.edu}}
\thanks{$^{4}${\tt\small zhigen.zhao@gatech.edu}}
\thanks{$^{5}${\tt\small yangning726@gmail.com}}
\thanks{$^{6}${\tt\small yunhao.liu@bytedance.com}}
}

\teaser{
  \centering
  \includegraphics[width=\textwidth]{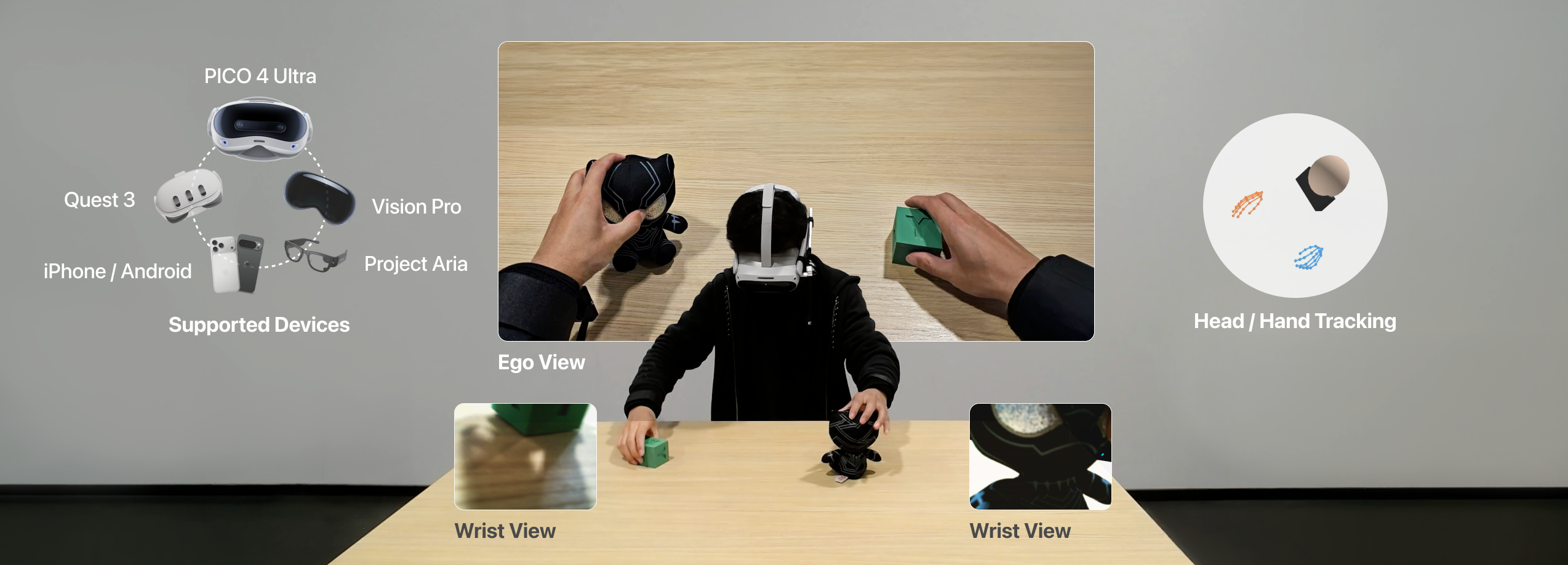}
  \captionof{figure}{\textbf{EgoKit-PICO use case.} EgoKit provides a unified egocentric data collection workflow across six types of devices---PICO 4 Ultra (EgoKit-PICO), Apple Vision Pro (EgoKit-AVP), Project Aria (EgoKit-Aria), iPhone (EgoKit-iOS), Android (EgoKit-Android), and Meta Quest 3 (EgoKit-Quest)---to capture ego-view and wrist-view video with off-the-shelf low-cost accessories. On headsets, such as PICO 4 Ultra and Meta Quest 3, it additionally logs 6 DoF head tracking and OpenXR-standard 26-joint hand tracking data.}%
  \label{fig:teaser}%
}

\maketitle
\thispagestyle{empty}
\pagestyle{empty}

\begin{abstract}
Egocentric video is increasingly used as a data source for robot learning, activity understanding, and embodied AI research, but collecting it at scale remains fragmented in practice: each candidate host device, such as an Android phone, iPhone, iPad, smart glasses, or extended reality (XR) headset, exposes a different SDK, a different policy on raw camera access, and different limitations on external USB cameras and on-device tracking. Synchronized ego-view and wrist-view capture is therefore typically obtained by either committing to a single proprietary platform or building one-off rigs that do not transfer across devices.
To address this gap, we present EgoKit, a toolkit that exposes the same egocentric recording workflow across six heterogeneous host devices. Across all supported devices, EgoKit presents the same recording interaction and produces locally stored video with a uniform log format; on XR headsets, it additionally logs head pose and OpenXR-standard 26-joint hand tracking aligned to the video streams. The companion accessories, including two wrist cameras with mounts, a head strap, and a USB-C hub, add wrist-view capture to any supported host without custom hardware fabrication. To validate that EgoKit data can train robot policies, we replicated HumanEgo with 41 demonstrations recorded by EgoKit-PICO: in simulation, policies trained on 20 or 40 of them placed a cube on a tray in 60--74\% of 100 unseen layouts, about as often as policies trained on Project Aria recordings of the same layouts. EgoKit is available at \url{https://egokit.chuange.org/}.
\end{abstract}

\section{Introduction}

\begin{figure*}[t]
    \centering
    \includegraphics[width=\linewidth]{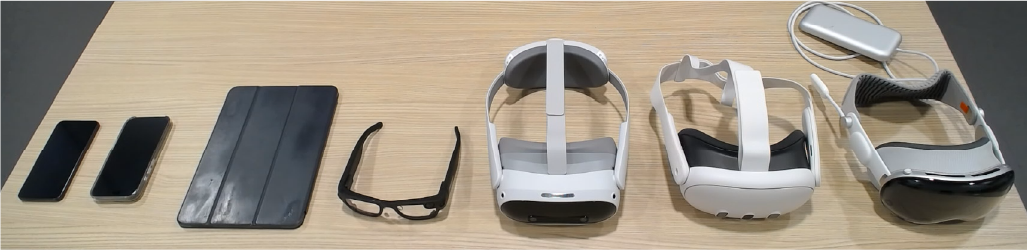}
    \caption{Devices supported by EgoKit. From left to right: Samsung Galaxy S23, iPhone 16 Pro, iPad Pro 2018, Project Aria (Gen 1), PICO 4 Ultra, Meta Quest 3, and Apple Vision Pro.}
    \label{fig:hardware}
\end{figure*}%

Scaling data has become a central driver of progress in robot learning, but collecting demonstrations directly on robots through teleoperation is expensive, hardware-dependent, and difficult to scale beyond a single lab. Egocentric human video has emerged as a complementary data source: humans perform manipulation tasks at much higher throughput than any robot fleet, and a growing body of datasets and methods has established that ego video can support meaningful transfer to robot policies. As these modeling efforts continue to push the demand for ego data upward, the practical question of how to actually record such data---at low cost, on the devices a team already owns---becomes increasingly pressing. Most teams already have at least some of the pieces: a spare Android phone or iPhone, a headset bought for development, a pair of smart glasses from a research grant, perhaps a few USB cameras in a drawer. What they lack is a recording system that fits these existing pieces. Off-the-shelf data collection products do not accept arbitrary host devices and require an additional hardware purchase that can run into the thousands of dollars; meanwhile, each candidate host platform exposes its own SDK, its own policy on raw camera access, and its own constraints on external USB cameras and on-device tracking, with no single workflow that spans them.

Several existing efforts have tackled parts of this problem. EgoVerse's phone-based pipeline~\cite{punamiya2026egoverse} head-straps an iPhone for 1080p ultrawide ego capture and reconstructs head and hand poses server-side, lowering the barrier to ego-only recording on a single device, but it does not address wrist-view capture, on-device tracking, or hosts beyond the iPhone. Research-grade platforms such as Project Aria~\cite{engel2023project} deliver high-quality ego streams with calibrated head and hand signals, but the hardware is not broadly available and is itself a single device. Commercial data collection products such as GenRobot DAS Ego~\cite{DASEgoGe56:online}, Sunday's Skill Capture Glove~\cite{SundayRo89:online}, PiKa~\cite{PIKA72:online}, and Lumos's FastUMI Ego~\cite{LumosUMI74:online} bundle ego and wrist capture into turnkey rigs, but each is tied to its own proprietary hardware stack and does not accept arbitrary host devices. On the teleoperation side, XRoboToolkit~\cite{zhao2026xrobotoolkit} demonstrates that an OpenXR-based abstraction can unify head, controller, and hand streams from PICO 4 Ultra and Meta Quest 3 into a common interface---but this work targets robot control, not data collection, and does not extend to phones or smart glasses. What is missing is a recording-side counterpart: a single workflow that spans phones, tablets, smart glasses, and XR headsets, supports wrist views as a first-class signal, and runs on hardware most teams already own.

We present EgoKit, a toolkit that fills this gap by exposing the same egocentric recording workflow across six heterogeneous host devices: Android phones, iPhones, Project Aria glasses, Apple Vision Pro, Meta Quest 3, and PICO 4 Ultra, with the iPad additionally supported as a wrist-view recorder, as shown in Fig.~\ref{fig:hardware}. Rather than designing new hardware, EgoKit treats each device as a host and ships a per-platform application that implements a shared recording workflow: operators trigger recording the same way on every device, each session produces locally stored video with a uniform log format, and---on XR headsets, where the platform supports it---head pose and 26-joint hand tracking are logged and aligned to the video streams. To extend any of these hosts with wrist-view capture, EgoKit pairs the software with consumer-grade accessories built entirely from off-the-shelf parts: two USB wrist cameras with mounts, a head strap, and a USB-C hub. The accessories cost about \$151, require no custom fabrication, and work across all supported devices. Together, these design choices let a team capture synchronized ego-view and wrist-view data on any of the supported hosts using a single operator workflow, without locking into any one vendor's stack.

The contributions of this work are as follows:

\begin{itemize}
    \item \textbf{A family of applications.} We release EgoKit-Android, EgoKit-iOS, EgoKit-iPad, EgoKit-Aria, EgoKit-AVP, EgoKit-Quest, and EgoKit-PICO, which bring the same egocentric recording workflow to every supported device, with head-tracking and OpenXR-standard 26-joint hand-tracking data logged on XR headsets.
    
    \item \textbf{A cross-platform accessory configuration.} Built from low-cost off-the-shelf consumer-grade parts, the accessories extend any of these hosts with two wrist-view cameras without custom hardware fabrication.
    
    \item \textbf{A consolidated record of platform-level constraints.} We identify which OS-level and SDK-level constraints actually bite in practice on each platform, covering raw camera access policies, external USB camera support, tracking API availability, and entitlement requirements.
    
    \item \textbf{Per-device measurements with selection guidance.} We report the resulting video and tracking characteristics measured per device, from which we derive practical guidance for teams selecting hardware for their own egocentric capture pipelines.

    \item \textbf{Validation for robot learning.} Replicating HumanEgo~\cite{humanego2026} in simulation, we show that demonstrations recorded with EgoKit-PICO train pick-and-place policies that succeed about as often as policies trained on recordings from Project Aria, the device HumanEgo was designed for.
\end{itemize}

\section{Related Work}

\begin{figure*}[!t]
    \centering
    \includegraphics[width=\linewidth]{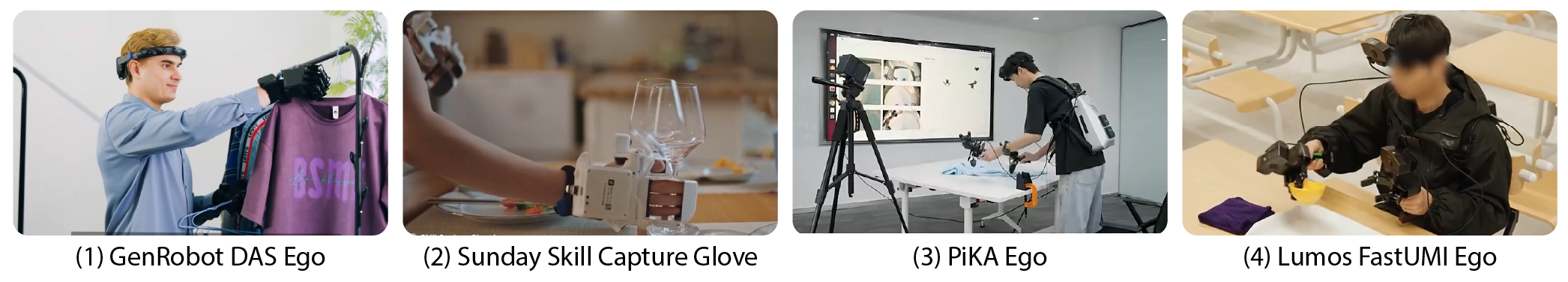}
    \caption{Examples of industry products for egocentric and/or wrist video data collection. Images are courtesy of the official product trailers.}
    \label{fig:other_hardware}
\end{figure*}

\subsection{Egocentric Data for Manipulation}

Simulation can sidestep physical deployment while preserving interactive supervision: Lucid-XR runs MuJoCo-class physics inside an XR headset and couples trajectories with language-steerable visual amplification, yielding end-effector-centric data that is portable across embodiments~\cite{ravan2026lucidxr}. At the opposite extreme, DemoBot shows that a single third-person RGB-D clip can seed contact-rich bimanual skills when dense hand--object estimates are treated as motion priors for residual reinforcement learning rather than pure imitation~\cite{xu2026demobot}. Between these poles sits a large family of methods that train vision-language-action models from human-centric video: EgoVLA predicts wrist and MANO-parameterized hand motion from egocentric frames and language, then bridges to robots via inverse kinematics and retargeting~\cite{yang2025egovla}; VITRA demonstrates automatic lifting of in-the-wild footage into short-horizon VLA episodes using reconstruction-based 3D hands, motion-based segmentation, and language labeling~\cite{li2025vitra}; CLAP aligns video transitions to discrete codes grounded in real robot trajectories so latent ``actions'' remain executable~\cite{zhang2026clap}; InternVLA-A1 mixes robot trajectories, simulation, and human video under a single understanding-generation-action architecture~\cite{cai2026internvla}; EgoScale scales ego pretraining and adds a small aligned human--robot mid-training stage with matched sensing~\cite{zheng2026egoscale}; EgoHumanoid targets whole-body loco-manipulation from portable ego capture plus limited robot teleoperation~\cite{shi2026egohumanoid}. Video-action modeling and world models argue that dynamics should not be learned only from static VLMs: mimic-video conditions a flow-based inverse-dynamics decoder on latent plans from a pretrained video backbone~\cite{pai2025mimicvideo}; DexWM predicts latent futures from keypoint-difference actions derived from ego human video and non-dexterous robot clips~\cite{goswami2026dexwm}; DreamDojo pretrains at tens of thousands of ego hours with continuous latent actions before adapting to robot controls~\cite{gao2026dreamdojo}. Synthetic scene generation (SAGE) complements ego video by producing simulator-valid 3D layouts and teacher policies without human recording~\cite{xia2026sage}. The corresponding rows of Table~\ref{tab:ego_related} summarize the sensing modality and stated collection philosophy of these entries in one place.

\subsection{Egocentric Data for Humanoid and Whole-Body Control}

Whole-body and mobile settings amplify both the value and the difficulty of ego sensing: HoMMI augments handheld manipulation interfaces with a head camera and global pose from commodity devices, then transfers via embodiment-aware visuals and relaxed gaze commands~\cite{xu2026hommi}; HumDex favors dense IMU-based body tracking and learned hand retargeting for humanoid teleoperation and optional human pretraining~\cite{heng2026humdex}; ActiveGlasses records stereo ego video and head motion from glasses and deploys an object-centric policy with a movable ``active'' camera arm~\cite{zou2026activeglasses}; MTC takes a VR-based route to whole-body locomotion data collection, recording the ego video an operator sees while making obstacle-avoidance decisions, alongside whole-body trajectories retargeted from immersive VR navigation~\cite{wang2026mtc}. Dexterous hardware heterogeneity motivates canonical hand URDFs and shared action spaces for cross-hand training~\cite{wei2026ohra}, while OmniXtreme addresses scalable dynamic humanoid tracking with generative pretraining plus execution-aware refinement~\cite{wang2026omnixtreme}. UltraDexGrasp illustrates large-scale purely synthetic bimanual grasp corpora when real contact data are scarce~\cite{yang2026ultradexgrasp}; OmniStream instead targets a unified streaming visual backbone for downstream VL and control~\cite{yan2026omnistream}. Egocentric whole-body pose from a single fisheye remains relevant as an upstream perception primitive for gloves-free capture~\cite{wang2023fisheyevit}. mimic-one is representative of high-fidelity robot-side dexterous logging via gloves or headset-driven teleoperation on a custom hand~\cite{nava2025mimicone}. These lines correspond to the locomotion, teleoperation, and representation rows in Table~\ref{tab:ego_related}.

\subsection{Egocentric and Embodied Datasets}

\begin{table*}[!t]
\small
\centering
\caption{Comparison of Egocentric Data Collection Systems and Datasets}
\label{tab:ego_related}
\renewcommand{\arraystretch}{1.22}
\begin{adjustbox}{max totalheight=0.92\textheight,center}
\begin{tabular}{p{3.2cm} p{3.5cm} p{4.1cm} p{4.9cm}}
\hline
\textbf{Work} & \textbf{Data Type} & \textbf{Hardware / Sensors} & \textbf{Main Contribution / Collection Method} \\
\hline

HoMMI~\cite{xu2026hommi} &
Whole-body mobile manipulation &
UMI gripper + iPhone head camera &
Captures coordinated locomotion and manipulation demonstrations with global head pose. \\

HumDex~\cite{heng2026humdex} &
Humanoid dexterous manipulation &
IMU suits + learned hand retargeting &
Simplified humanoid teleoperation pipeline with optional human pretraining. \\

ActiveGlasses~\cite{zou2026activeglasses} &
Egocentric manipulation demonstrations &
XREAL smart glasses + ZED stereo &
Stereo egocentric video and head motion capture with an active-vision camera arm. \\

HOCap~\cite{wang2025hocap} &
Hand--object capture &
Wearable cameras + external tracking &
Capture system producing dense 3D hand--object interaction annotations. \\

EgoLive~\cite{li2026egolive} &
Large-scale egocentric dataset &
Wearable first-person cameras &
1{,}680\,h stereo ego @ 60 FPS; real-world human task routines. \\

EgoVerse~\cite{punamiya2026egoverse} &
Consortium-scale egocentric demonstrations &
Project Aria (academic), custom rigs (industry), head-strapped iPhone (community) &
1{,}362\,h, 1{,}965 tasks, 240 scenes, 2{,}087 demonstrators; unified 3D hand and head poses via EgoDB; cross-lab human-to-robot transfer study on three robots. \\

HumanNet~\cite{deng2026humannet} &
First- and third-person human-centric video &
Internet-sourced video; no bespoke capture rig &
1M-hour corpus with systematic curation and interaction-centric annotations (captions, motion descriptions, hand/body signals) for embodied pretraining. \\

EgoScale~\cite{zheng2026egoscale} &
Wild egocentric manipulation video &
Curated wild ego video; aligned Vive + Manus + R1Pro &
20\,k\,h wild ego corpus with a two-stage human pretrain plus aligned mid-train methodology. \\

HOT3D~\cite{banerjee2024hot3d} &
3D hand--object interaction &
Project Aria + Quest~3 &
Multi-view egocentric RGB, depth, IMU, eye tracking, and hand--object ground truth. \\

Assembly101~\cite{sener2022assembly101} &
Assembly task dataset &
8 static + 4 ego cameras &
4{,}321 sequences over 101 toy assemblies; long procedural activities with dense temporal labels. \\

EgoDex~\cite{egodex} &
Egocentric dexterous manipulation &
Wearable cameras &
Dexterous hand--object interaction dataset from first-person views. \\

Gen2 Pilot~\cite{kong2025ariagen2pilot} &
Embodied AI dataset &
Project Aria Gen 2 sensors &
Egocentric demonstrations with spatial tracking and multimodal sensing. \\

UltraDexGrasp~\cite{yang2026ultradexgrasp} &
Synthetic bimanual grasp data &
Simulation &
Large-scale synthetic bimanual dexterous grasping corpus. \\

\hline
\end{tabular}
\end{adjustbox}
\end{table*}

Public corpora anchor scaling claims and evaluation: EgoLive releases long stereo ego video at high resolution with rich geometric and semantic annotations for manipulation-oriented tasks~\cite{li2026egolive}; HumanNet provides a one-million-hour human-centric video corpus spanning first- and third-person perspectives, paired with a systematic curation paradigm and interaction-centric annotations (captions, motion descriptions, hand and body signals) that enable motion-aware and interaction-aware pretraining for embodied agents~\cite{deng2026humannet}; EgoVerse takes the orthogonal stance of a continuously growing, consortium-scale ``living dataset'' rather than a static release, aggregating 1{,}362\,h of egocentric demonstrations across 1{,}965 tasks, 240 scenes, and 2{,}087 demonstrators from academic labs (Project~Aria, EgoVerse-A), industry partners (custom rigs, EgoVerse-I), and a community-accessible phone-based capture pipeline, all unified into 3D hand poses, head-pose, and language-annotated episodes via the EgoDB cloud system, and validated with a cross-lab human-to-robot transfer study on three robot platforms~\cite{punamiya2026egoverse}; HOT3D provides multi-view ego streams from Project Aria and Quest~3 with accurate hand--object ground truth under controlled capture~\cite{banerjee2024hot3d}; Assembly101 mixes static and ego viewpoints over long procedural assembly with dense temporal labels~\cite{sener2022assembly101}. Open X-Embodiment aggregates over a million real robot trajectories across many platforms for cross-embodiment policy research~\cite{openx2023embodiment}. Community hubs such as EgoDex~\cite{egodex}, Harmony4D~\cite{harmony4d}, Gen2 Pilot~\cite{kong2025ariagen2pilot}, HOCap~\cite{wang2025hocap}, and Meta's UmeTrack~\cite{han2022umetrack} broaden coverage of hand pose, pilot environments, and tracking tooling; readers should verify licensing and task overlap before merging splits. The dataset rows of Table~\ref{tab:ego_related} align with this subsection.

\subsection{Hardware and Capture Platforms}

Commercial hardware lowers friction for egocentric data collection, such as GenRobot's DAS Ego (Fig.~\ref{fig:other_hardware}(1))~\cite{DASEgoGe56:online}, Sunday's Skill Capture Glove (Fig.~\ref{fig:other_hardware}(2))~\cite{SundayRo89:online}, PiKa's setups (Fig.~\ref{fig:other_hardware}(3))~\cite{PIKA72:online}, and Lumos's FastUMI Ego (Fig.~\ref{fig:other_hardware}(4))~\cite{LumosUMI74:online}.

EgoVerse's phone-based capture path is a closely related point of reference: an iPhone is head-strapped to record 1080p\,/\,30\,fps ultrawide ego video, with 6-DoF head pose recovered by cloud-side visual tracking and 21-keypoint per-hand poses estimated server-side~\cite{punamiya2026egoverse}. EgoKit covers the same single-device ego-only use case while additionally supporting wrist-view cameras and on-device head/hand tracking on XR headsets, and broadening the host envelope to Android, iPad, smart glasses, Apple Vision Pro, Meta Quest~3, and PICO 4 Ultra. These platforms do not replace dataset papers but clarify how supervision is acquired in production pilots; Table~\ref{tab:ego_related} consolidates representative hardware and capture pipelines for quick comparison. For conceptual ordering of transfer mechanisms and dataset trends, the survey of Ma et al.~\cite{ma2026survey} remains the broadest guide.

\section{EgoKit Toolkit}

\subsection{Overview}\label{sec3-1}

\begin{figure*}[t]
    \centering
    \includegraphics[width=\linewidth]{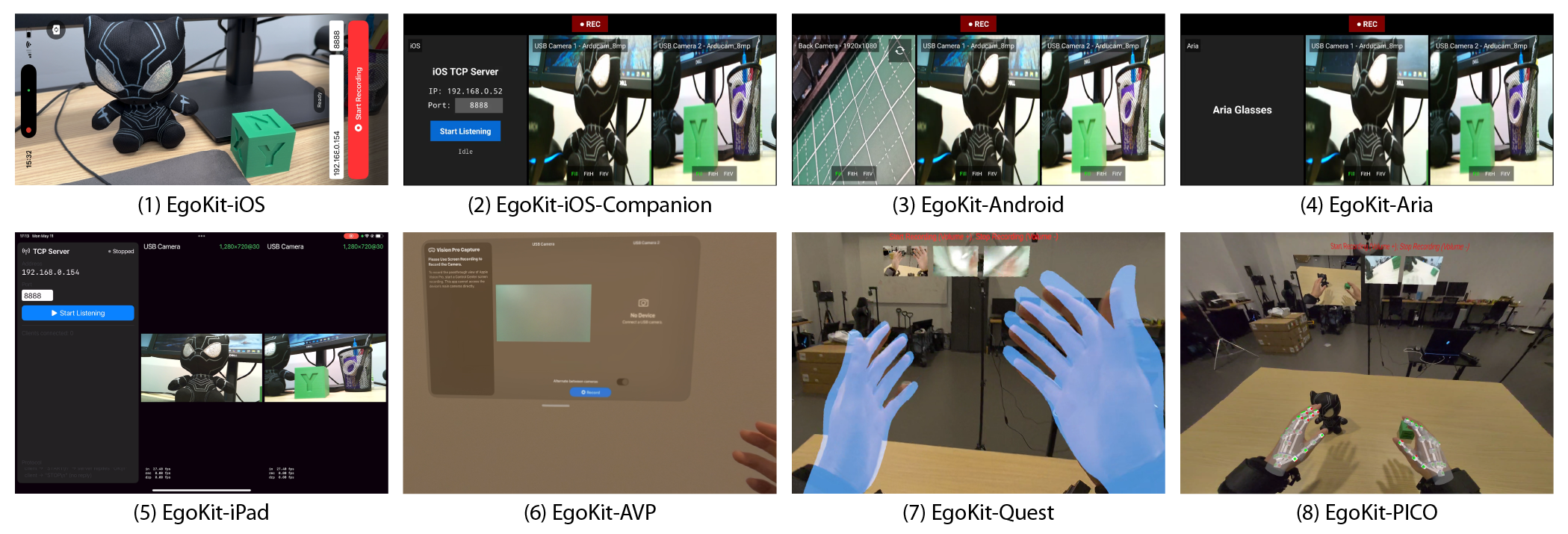}
    \caption{User Interface of the EgoKit family. Please refer to Section~\ref{sec3-1} for their explanations.} 
    \label{fig:interface}
\end{figure*}

The user interface for each device is shown in Fig.~\ref{fig:interface}, where the label indicates the project name. Fig.~\ref{fig:interface}(1) is the interface running on an iPhone, and Fig.~\ref{fig:interface}(2) is the interface running on an Android phone. Fig.~\ref{fig:interface}(1) captures only the ego view using the head mount, and Fig.~\ref{fig:interface}(2) is its counterpart that captures the wrist view. Fig.~\ref{fig:interface}(3) is the application running on an Android phone that can work alone. Fig.~\ref{fig:interface}(4) is the application running on an Android phone to capture the wrist view, while the ego view comes from the Project Aria glasses connected to and controlled by the same Android phone. Fig.~\ref{fig:interface}(5) is the interface running on an iPad to capture the wrist view, which can be controlled by Fig.~\ref{fig:interface}(1). Fig.~\ref{fig:interface}(6) shows the interface on Apple Vision Pro. Even though two wrist cameras are supported, only one is connected and previewing, because Apple Vision Pro does not support recording two USB cameras simultaneously. As raw camera access requires an entitlement, the ego view recording is substituted by its internal view recording. The interface disappears when recording starts. The recording stops and the UI reappears when the Digital Crown button is pressed. Fig.~\ref{fig:interface}(7) is the interface running on the Meta Quest 3 headset, where three preview windows show the ego view and wrist views, along with the hand models. The camera previews and hand models disappear when recording is started by pressing the Volume + button, and reappear when recording is stopped by pressing the Volume - button. Fig.~\ref{fig:interface}(8) is the interface running on the PICO 4 Ultra headset, where the ego view, two wrist views, and hand skeletons are shown. They disappear when the Volume + button is pressed to start recording, and reappear when the Volume - button is pressed to stop recording. The recording is automatically saved when the stop recording action is triggered.

\subsection{Design Choices}

We aim to devise a unified interface for egocentric data collection. At the same time, the capability differences across devices must be considered. For example, the Android platform is much more open than the iOS platform, and headsets offer more features than phones, such as hand tracking. Moreover, the differences among SDKs introduce additional constraints; for instance, the smart glasses supported in EgoKit lack a mobile SDK. The interaction should also be simple and intuitive enough for the data collection operator. In this context, the host device refers to the device running the application, either a phone or a headset.


\begin{table}[t]
\centering
\caption{Host devices used to run the EgoKit applications.}
\label{tab:hardware_hosts}
\footnotesize
\begin{tabularx}{\columnwidth}{@{}Xl@{}}
\toprule
\textbf{Device} & \textbf{OS Version} \\
\midrule
Samsung Galaxy S23           & Android 15 \\
Project Aria Glasses$^{1}$   & Glasses OS 4975080.310.70 \\
iPhone 16 Pro                & iOS 26.4.2 \\
iPad Pro 2018 (11-inch)      & iPadOS 18.7.1 \\
Apple Vision Pro$^{2}$       & visionOS 26.3 \\
Meta Quest 3$^{3}$           & Horizon OS 85.0 \\
PICO 4 Ultra                 & PICO OS 5.15.4.U \\
\bottomrule
\end{tabularx}
\\[2pt]
\begin{flushleft}\footnotesize
$^{1}$Gen 1. Aria App on Android 265.1.0.13.0.\\
$^{2}$Requires Developer Strap (\$299).\\
$^{3}$Horizon OS 2.x is not supported.
\end{flushleft}
\end{table}

\ssection{Simple Consistent Interaction.} The main interactions include opening the application and starting/stopping recordings. For phone applications, it can use a simple touch to open. For headset applications, it can use a pinch to open. Specifically, on Android-based platforms, such as a regular Android phone, a Meta Quest headset, and a PICO headset, the application can auto-open when the wrist cameras are connected. This can reduce the interaction burden. Moreover, for the host device, we use volume keys as shortcuts to start/stop recordings. Specifically, the volume \textbf{+} key is to start recording, and the volume \textbf{-} key is to stop recording. The recording will be saved automatically once the stop recording action is triggered.

\ssection{Local Storage.} Recordings should be saved locally and named consistently. Streaming elsewhere requires a network connection, consumes more power, and is less reliable. Video recordings are saved as MP4 files using H.264 encoding, and other data (e.g., metadata and pose data) are saved as plain text for cross-platform compatibility.

\ssection{Unleashing Hardware Potential.} Hardware-specific features should be fully used. For example, a hardware encoder can accelerate video processing, and headsets support head tracking and hand tracking.

\subsection{Hardware and Accessories}

\begin{figure}[t]
    \centering
    \includegraphics[width=\linewidth]{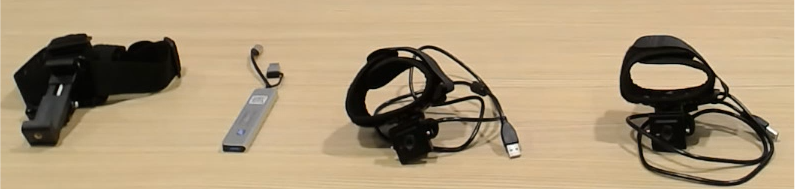}
    \caption{Off-the-shelf consumer-grade accessories. From left to right: head strap, USB-C hub, and two USB cameras with the wrist mount.}
    \label{fig:accessories}
\end{figure}

Table~\ref{tab:hardware_hosts} details the host devices that run the EgoKit applications. Table~\ref{tab:accessories} details the accessories used for wrist view recording. Fig.~\ref{fig:accessories} illustrates the accessories.


\begin{table}[!t]
\centering
\caption{Accessories used for wrist view recording. Superscripts on items are clickable links to the product page.}
\label{tab:accessories}
\footnotesize
\begin{tabularx}{\columnwidth}{@{}Xr@{}}
\toprule
\textbf{Item} & \textbf{Cost} \\
\midrule
Wrist camera (\(\times\)2)\,\href{https://www.amazon.com/dp/B09BR1RNSN}{$^{\text{[link]}}$} & \textasciitilde\$80 \\
Wrist camera mount (\(\times\)2)\,\href{https://www.amazon.com/dp/B07H4MSW9Q}{$^{\text{[link]}}$} & \textasciitilde\$30 \\
Head strap\,\href{https://www.amazon.com/dp/B08R72LPNQ}{$^{\text{[link]}}$} & \textasciitilde\$20 \\
USB-C hub & \textasciitilde\$20 \\
Machine screw (\(\times\)2)\,\href{https://www.homedepot.com/p/317479058}{$^{\text{[link]}}$} & \textasciitilde\$1 \\
\midrule
\textbf{Total} & \textbf{\textasciitilde\$151} \\
\bottomrule
\end{tabularx}
\end{table}

\subsection{EgoKit Family}

\subsubsection{EgoKit-Android}

EgoKit-Android is a Kotlin application with a Jetpack Compose UI. The on-device ego view is captured through the \texttt{Camera2} API at 1920$\times$1080, while up to two wrist cameras are streamed through the AndroidUSBCamera library. All streams are encoded in hardware with \texttt{MediaCodec} and muxed to H.264 MP4 by \texttt{MediaMuxer}: the internal camera uses Surface-mode input from the Camera2 capture session, and the USB cameras feed NV21 frames into a byte-buffer encoder. \texttt{dispatchKeyEvent} maps Volume\,$+$/$-$ to start/stop. Recordings are written to \texttt{DCIM/\bk Recordings/\bk \{timestamp\}/} with per-camera MP4 files and a \texttt{log.txt} listing per-frame timestamps for post-hoc alignment. Listing~\ref{android-log-example} shows an example.

\begin{lstlisting}[label=android-log-example, caption=EgoKit-Android log file example]
Recording Session: 20260511_162515
================================

File: internal.mp4
  Source: Internal Camera
  Total frames: 763
  First frame timestamp: 1778531115762 ms
  Last frame timestamp: 1778531141168 ms
  Duration: 25406 ms

File: usb2.mp4
  Source: USB Camera 2
  Total frames: 483
  First frame timestamp: 1778531115740 ms
  Last frame timestamp: 1778531141551 ms
  Duration: 25811 ms

File: usb1.mp4
  Source: USB Camera 1
  Total frames: 473
  First frame timestamp: 1778531115784 ms
  Last frame timestamp: 1778531141589 ms
  Duration: 25805 ms
\end{lstlisting}

\subsubsection{EgoKit-Aria}

Project Aria has no public mobile SDK, so EgoKit-Aria operates in a fully decoupled fashion: the glasses record the ego view on-device under profile~18 (1408$\times$1408 fisheye VRS at 10\,fps), and a paired Android phone captures the wrist cameras. Because there is no programmatic SDK to start/stop Aria recording, EgoKit-Aria drives the official Aria companion app through an \texttt{AccessibilityService} that locates and clicks UI elements by text, allowing the host phone to trigger glasses recording in lockstep with its own. The wrist-camera pipeline is identical to EgoKit-Android, and Aria's egocentric video is reconciled with the wrist clips offline using the per-session \texttt{log.txt} timestamps. Listing~\ref{vrs-to-mp4} illustrates how to convert a vrs file to an mp4 file.

\begin{lstlisting}[language=Python, label=vrs-to-mp4, caption=Python script to convert vrs to mp4]
# `pip install projectaria-tools` first
# Note: this script requires the ffmpeg module.

from projectaria_tools.utils.vrs_to_mp4_utils import convert_vrs_to_mp4

input_vrs = "input.vrs"
output_mp4 = "output.mp4"
 
convert_vrs_to_mp4(input_vrs, output_mp4)
\end{lstlisting}

\subsubsection{EgoKit-iOS}

iOS\,26 prevents an iPhone from simultaneously driving its built-in camera and an external USB device, so EgoKit splits ego- and wrist-view capture across two paired devices. The iPhone app (EgoKit-iOS) records the ego view with \texttt{AVCaptureSession} on the built-in wide-angle camera and writes H.264 \texttt{.mov} via \texttt{AVAssetWriter}. Hardware volume keys are intercepted without changing the system volume. The companion app (EgoKit-iOS-Companion) runs on a tethered Android phone and accepts START/STOP commands over a TCP socket built on Apple's \texttt{Network} framework; the iPhone waits for an \texttt{OK} acknowledgement before beginning local capture so both devices start within one round-trip. The companion reuses the EgoKit-Android stack and writes recordings into a timestamped session folder. No NTP synchronization is performed; the session-folder timestamps and per-camera \texttt{log.txt} are used to align streams in post-processing.

\subsubsection{EgoKit-iPad}

The iPad variant pairs the same EgoKit-iOS iPhone app with EgoKit-iPad, an iPadOS\,18.7.1 application that exploits iPadOS's first-class external-camera support to host both wrist cameras on a single device. Cameras are discovered via \texttt{AVCaptureDevice.\bk DiscoverySession} filtered on the \texttt{.external} device type, with hot-plug events handled through \texttt{was\bk Connected\bk Notification} / \texttt{was\bk Disconnected\bk Notification}. When the hardware permits it, an \texttt{AV\bk Capture\bk MultiCam\bk Session} streams both cameras concurrently; format negotiation prefers 1280$\times$720@30 and falls back to the closest supported descriptor. Sample buffers are split between an \texttt{AVSampleBuffer\bk DisplayLayer} preview and an \texttt{AVAssetWriter} writing H.264 in a \texttt{.mov} container, with PTS timestamps rebased from the capture-session clock for synchronization. Recording start/stop is driven from the iPhone over a TCP socket implemented with the \texttt{Network} framework (\texttt{TCPServer.swift}); the resulting clips are saved to the Photos library.

\subsubsection{EgoKit-AVP}

EgoKit-AVP is an application targeting visionOS\,26 that can record from the user-attached wrist USB camera without the enterprise-license entitlements required to access Vision Pro's own scene cameras. Because visionOS allows only a single active USB input at a time, the \texttt{CaptureManager} actor implements a soft round-robin: when two cameras are connected, it swaps the \texttt{AVCaptureSession} input between them at roughly 100\,Hz, exposing both as live \texttt{AVSampleBuffer\bk DisplayLayer} previews while only one stream is recorded at any instant. External cameras are discovered through \texttt{AVCaptureDevice.\bk DiscoverySession} with the \texttt{.external} device type. The recorder uses \texttt{AVAssetWriter} with HEVC at 1280$\times$720, writing \texttt{.mov} files saved to the Photos library via \texttt{PHPhotoLibrary}. The current implementation does not include head- and hand-tracking annotations.

\subsubsection{EgoKit-Quest}

EgoKit-Quest is a Unity application on Quest\,3, built on the Meta XR SDK\,v81 with the Oculus XR provider for head-pose data and \texttt{OVRSkeleton}/Meta Hand Tracking for 26-joint per-hand poses in OpenXR joint order. Recording is delegated to a custom native plugin (\texttt{libxrcorehelper.aar}) that wires Camera2 directly to a \texttt{MediaCodec} input Surface for hardware H.264 encoding of the front-facing passthrough camera (1280$\times$720@\textasciitilde{}60), gated by the \texttt{horizonos.\bk permission.\bk USB\_\bk CAMERA} permission. Wrist cameras share the same plugin via an NV21 preview-data path at 1280$\times$720@\textasciitilde{}27. Pose logs (head plus 52 hand joints, 53 transforms per frame) are timestamped to 1\,ms and written from a background thread into \texttt{poses.txt} alongside \texttt{internal.mp4}/\texttt{usb1.mp4}/\texttt{usb2.mp4} under \texttt{DCIM/\bk Recordings/\bk \{timestamp\}/}. Volume\,$+$/$-$ is mapped to start/stop recording.

\subsubsection{EgoKit-PICO}

EgoKit-PICO is a Unity application on the PICO\,4 Ultra, built on the PICO Unity Integration SDK with the \texttt{PXR\_Loader} backend (rather than OpenXR) so it can call \texttt{PXR\_\bk Manager.\bk EnableVideoSeeThrough} and the proprietary \texttt{PXR\_CameraImage} API to access the headset's see-through cameras directly. The high ego-view rate (1280$\times$960 at \textasciitilde{}89\,fps) reflects the device's native passthrough capability. Encoding reuses the shared XRCoreHelper native plugin (\texttt{libxrcorehelper.aar}) to bridge Camera2 output to a \texttt{MediaCodec} input Surface for hardware H.264 at \textasciitilde{}8\,Mbps. Wrist cameras are routed through the same plugin's NV21 preview-data path and recorded as \texttt{usb1.mp4}/\texttt{usb2.mp4} at 1280$\times$720@\textasciitilde{}27. Hand tracking uses Unity's \texttt{com.\bk unity.\bk xr.\bk hands} subsystem (which binds to the PICO XRHandSubsystem) and yields 26 joints per hand in OpenXR order; head and hand transforms are sampled at the main-loop rate into a lock-free ring buffer and serialized to \texttt{poses.txt} from a background thread. Volume\,$+$/$-$ is mapped to start/stop recording. Listing~\ref{pico-log-example} shows a log file example. Listing~\ref{poses-schema} shows the schema of the poses file.

\begin{lstlisting}[label=pico-log-example, caption=EgoKit-PICO log file example]
Recording Session: 20260512_044648
Started:  2026-05-12 04:46:48.572
Ended:    2026-05-12 04:47:18.443
================================

File: internal.mp4
  Source: Internal Camera
  Total frames: 2623
  First frame timestamp (unix ms): 1778528808577
  Last frame timestamp  (unix ms): 1778528838419
  Duration (ms): 29842

File: usb1.mp4
  Source: USB Camera 1
  Total frames: 600
  First frame timestamp (unix ms): 1778528808675
  Last frame timestamp  (unix ms): 1778528838294
  Duration (ms): 29619

File: usb2.mp4
  Source: USB Camera 2
  Total frames: 540
  First frame timestamp (unix ms): 1778528808725
  Last frame timestamp  (unix ms): 1778528838357
  Duration (ms): 29632

File: poses.txt
  Source: Poses
  Total frames: 2622
  First frame timestamp (unix ms): 1778528808795
  Last frame timestamp  (unix ms): 1778528838153
  Duration (ms): 29358
\end{lstlisting}

\begin{lstlisting}[label=poses-schema, caption=Schema of the poses file.]
# idx t_ms head_pos(x y z) head_rot(x y z w) left[26]:(pos+rot) right[26]:(pos+rot)
# joints in OpenXR XR_HAND_JOINT order: 0=Palm 1=Wrist 2..5=Thumb(M,P,D,T) 6..10=Index(M,P,I,D,T) 11..15=Middle 16..20=Ring 21..25=Little
\end{lstlisting}

\section{Use Cases}

\begin{figure*}[t]
    \centering
    \includegraphics[width=\linewidth]{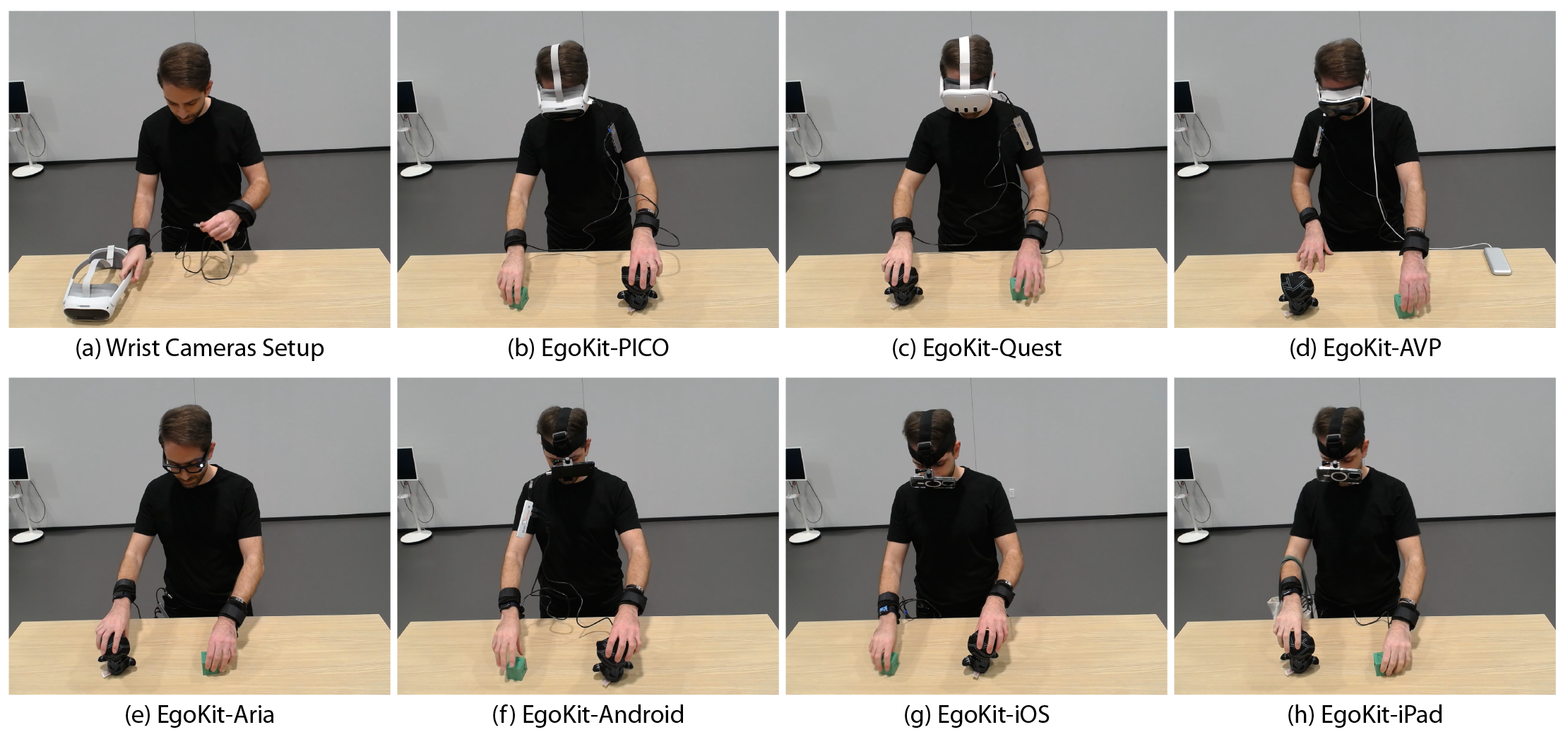}
    \caption{Various setups of EgoKit. Please refer to Section~\ref{sec4-1} for their explanations.}
    \label{fig:setups}
\end{figure*}

\begin{figure*}[t]
    \centering
    \includegraphics[width=\linewidth]{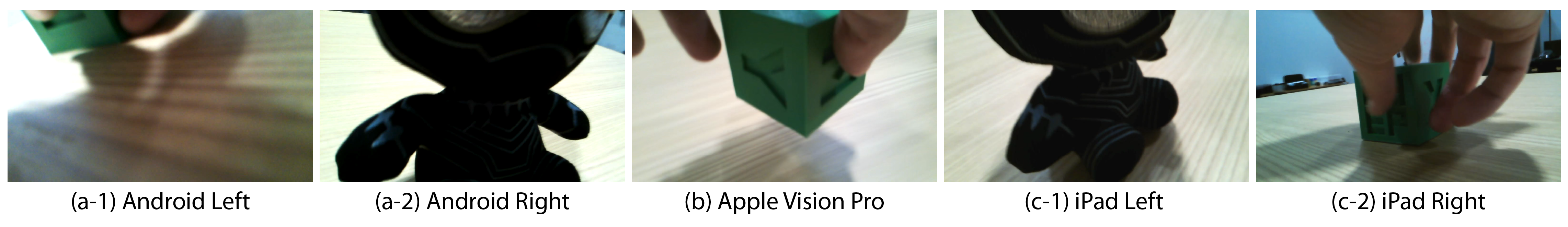}
    \caption{Frame examples of wrist view recordings. (a-1) and (a-2) are from the setup where a USB-C hub is connected to an Android device (headset or phone). (b) is from Apple Vision Pro. (c-1) and (c-2) are from the setup where a USB-C hub is connected to an iPad.}
    \label{fig:wristviews}
\end{figure*}

\begin{figure*}[t]
    \centering
    \includegraphics[width=\linewidth]{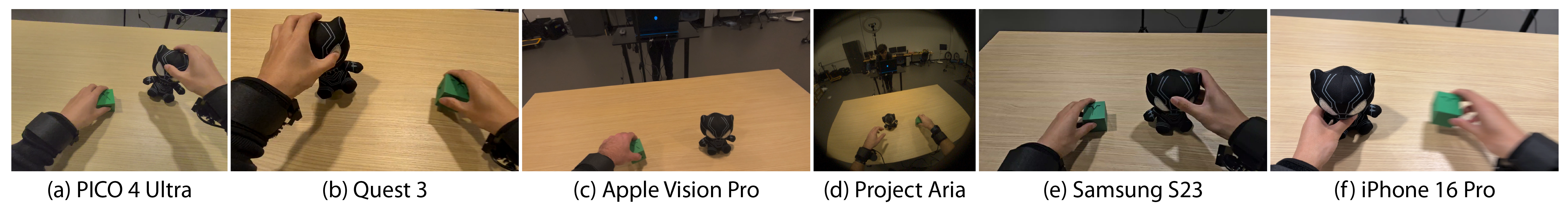}
    \caption{Frame examples of egocentric video recordings using different devices. The label indicates the host device that records the egocentric view. Images are scaled to the same height while keeping their aspect ratio for a compact and clear layout.}
    \label{fig:egoviews}
\end{figure*}

In this section, we demonstrate the use cases of EgoKit.

\subsection{Setup}\label{sec4-1}
Figure~\ref{fig:setups} demonstrates the use cases with different devices. Each USB camera is mounted on a wrist and then connected to the USB-C hub. After the two USB cameras are mounted, the USB-C hub can be plugged into the host device, as shown in Fig.~\ref{fig:setups}(a). This setup applies to the PICO 4 Ultra headset (Fig.~\ref{fig:setups}(b)), the Meta Quest 3 headset (Fig.~\ref{fig:setups}(c)), and the Apple Vision Pro headset (Fig.~\ref{fig:setups}(d)). Note that, because Apple Vision Pro does not support two USB cameras simultaneously, only one USB camera is connected to the USB-C hub.

Since the Android phone can support two USB cameras, the USB-C hub is connected to the Android phone to record wrist views. This setup applies to Project Aria (Fig.~\ref{fig:setups}(e)), the Android phone (Fig.~\ref{fig:setups}(f)), and the iPhone (Fig.~\ref{fig:setups}(g)). The iPhone uses this setup because it does not officially support USB cameras, even in iOS 26.4.2, whereas the iPad supports two USB cameras even in iPadOS 18.7.1 (Fig.~\ref{fig:setups}(h)).

The iPhone can record egocentric views, but it needs another device to record wrist views, either an Android phone (Fig.~\ref{fig:setups}(g)) or an iPad (Fig.~\ref{fig:setups}(h)).

\subsection{Ego Views}

Figure~\ref{fig:egoviews} illustrates some frame examples of recorded egocentric videos. The label is the host device that records the egocentric view.

\subsection{Wrist Views}

\begin{table*}[!ht]
\centering
\caption{Comparison of various setups for ego-view and wrist-view recording, and head/hand tracking capabilities across host devices. Head/Hand Tracking: \cmark = supported, \xmark = not supported, M = manual (post-hoc) reconstruction.}
\label{tab:device_comparison}
\footnotesize
\begin{adjustbox}{width=\textwidth}
\begin{tabular}{llcccccccccc}
\toprule
\multirow{2}{*}{\textbf{Host Device}} & \multirow{2}{*}{\textbf{Edge Device}} & \multicolumn{4}{c}{\textbf{Ego View}} & \multicolumn{4}{c}{\textbf{Wrist View}} & \multirow{2}{*}{\textbf{\shortstack{Head\\Tracking}}} & \multirow{2}{*}{\textbf{\shortstack{Hand\\Tracking}}} \\
\cmidrule(lr){3-6} \cmidrule(lr){7-10}
 &  & Format & Resolution & FPS & Bitrate & Format & Resolution & FPS & Bitrate &  &  \\
\midrule
Android              & --      & mp4 & 1920$\times$1080 & 29.84 & 4.02 Mbps  & mp4 & 1280$\times$720 & 30.00 & 1.98 Mbps  & \xmark & \xmark \\
Project Aria$^{1}$ & Android & vrs & 1408$\times$1408 & 10.00 & 8.85 Mbps  & mp4 & 1280$\times$720 & 30.00 & 1.98 Mbps  & M & M \\
iPhone               & Android & mov & 1920$\times$1080 & 30.00 & 15.46 Mbps & mp4 & 1280$\times$720 & 30.00 & 1.99 Mbps  & \xmark & \xmark \\
iPhone               & iPad    & mov & 1920$\times$1080 & 30.00 & 15.44 Mbps & mov & 1280$\times$720 & 27.37 & 6.13 Mbps  & \xmark & \xmark \\
Apple Vision Pro$^{2}$ & --  & mov & 1920$\times$1080 & 29.88 & 13.02 Mbps & mov & 1280$\times$720 & 27.38 (0.53)$^{\dagger}$ & 2.47 Mbps (51 kbps)$^{\dagger}$ & \cmark$^{\ddagger}$ & \cmark$^{\ddagger}$ \\
Meta Quest 3         & --      & mp4 & 1280$\times$720  & 59.42 & 9.04 Mbps  & mp4 & 1280$\times$720 & 27.37 & 3.75 Mbps  & \cmark & \cmark \\
PICO 4 Ultra$^{3}$         & --      & mp4 & 1280$\times$960  & 89.31 & 8.45 Mbps  & mp4 & 1280$\times$720 & 27.36 & 3.56 Mbps  & \cmark & \cmark \\
\bottomrule
\end{tabular}
\end{adjustbox}
\begin{flushleft}\footnotesize
$^{1}$Project Aria uses Profile 18 for recording.\\
$^{2}$Apple Vision Pro only supports a single wrist camera.\\
$^{\dagger}$In $XX (YY)$, $XX$ means for a single wrist camera, while $YY$ means for two wrist cameras by constant switching. $^{\ddagger}$Not included.\\
$^{3}$PICO 4 Ultra supports stereo RGB camera access. Here, only the left RGB camera is used.
\end{flushleft}
\end{table*}

Figure~\ref{fig:wristviews} illustrates some frame examples of recorded wrist view videos. Fig.~\ref{fig:wristviews}(a-1) and Fig.~\ref{fig:wristviews}(a-2) are from the setup where the USB-C hub is connected to an Android device (headset or phone), which includes EgoKit-PICO, EgoKit-Quest, EgoKit-Aria, EgoKit-Android, and EgoKit-iOS. Fig.~\ref{fig:wristviews}(b) is from the setup where the USB-C hub is connected to Apple Vision Pro using the Developer Strap. Note that it includes only one view, because Apple Vision Pro does not support recording two USB cameras. Fig.~\ref{fig:wristviews}(c-1) and Fig.~\ref{fig:wristviews}(c-2) are from the setup where the USB-C hub is connected to an iPad. It is worth mentioning that iPadOS 18.7.1 supports two USB cameras.

Table~\ref{tab:device_comparison} details the comparison across different setups. The host device refers to the device that records the egocentric view, while the edge device refers to the device that records the wrist view. If the edge device is not specified, the wrist cameras connect directly to the host device via a USB-C hub.

\section{Validating EgoKit Data for Robot Learning}\label{sec:policy}

\begin{figure*}[t]
    \centering
    \includegraphics[width=\linewidth]{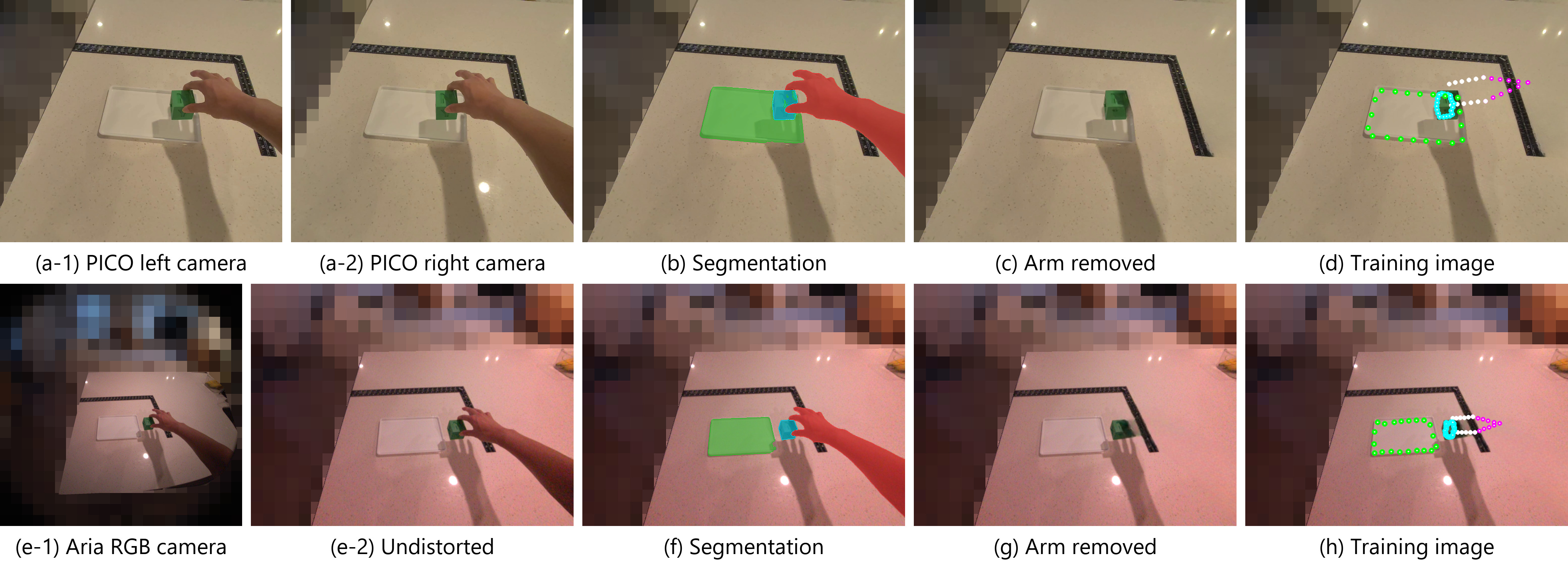}
    \caption{One frame of the same layout, recorded by both devices while the cube is carried, and HumanEgo's preprocessing of it. Top: PICO 4 Ultra with EgoKit-PICO; bottom: Project Aria (Gen 1). (a-1, a-2) The PICO's two see-through cameras, recorded side by side: the pipeline uses the left image and locates the objects by stereo matching between the two. (e-1) The Aria's fisheye RGB camera and (e-2) the same image undistorted for the pipeline. (b, f) Masks of the arm (red), the cube (cyan), and the tray (green). (c, g) The arm removed by inpainting. (d, h) The training image: a virtual gripper at the tracked hand's pose, drawn closed (white to magenta), and the keypoints of the cube and the tray. Everything but the countertop is pixelated.}
    \label{fig:humanego_data}
\end{figure*}

To validate that data recorded with EgoKit can train robot policies, we replicated HumanEgo~\cite{humanego2026}, which learns manipulation policies from minutes of human egocentric video, and trained it on demonstrations recorded with EgoKit-PICO. HumanEgo was designed for Project Aria (Gen 1) glasses and Meta's Machine Perception Services (MPS), so we recorded every layout with Aria glasses as well, as a reference. A simulated arm takes the place of the real robot.

\subsection{Method}

\ssection{HumanEgo.} HumanEgo reduces each recording to the trajectory of the wearer's hand and the 3D keypoints of the task objects. Its preprocessing segments the objects with open-vocabulary detection, tracks keypoints on them, and triangulates the keypoints; it then removes the arm from the images by inpainting and draws a virtual gripper at the hand and the keypoints on the objects (Fig.~\ref{fig:humanego_data}). A flow-matching policy takes such an image, together with tokens that describe the hand and the objects, and predicts the hand's 6-DoF pose and grasp state over the next 50 frames, which the robot's gripper follows in a closed loop. Our replication reproduced the authors' published preprocessing output on two of their released recordings, with object positions within 1\,mm.

\ssection{EgoKit-PICO as a Source.} We added an adapter that reads EgoKit-PICO recordings. The headset's own head pose and 26-joint hand tracking take the place of MPS, and the object keypoints are triangulated by stereo matching between the headset's two see-through cameras (Fig.~\ref{fig:humanego_data}(a-1, a-2)), whereas the single-camera Aria triangulates them from the head's motion as the wearer walks up to the table. The remaining stages are shared by both devices.

\ssection{Simulation.} In place of the authors' robot, HumanEgo's own inference loop drives a Trossen WidowX AI arm from MuJoCo Menagerie~\cite{zakka2022menagerie}, simulated in MuJoCo~\cite{todorov2012mujoco}. The simulator renders the camera image, provides the object masks, and removes the arm by rendering the scene without it (Fig.~\ref{fig:humanego_sim}(a, f)). With the authors' released policy, the simulated arm put the loaf on the plate in 47 of the 61 layouts recorded for their serve-bread task (77\,\%).

\subsection{Task and Data}

The task is a one-handed pick-and-place: put a 5\,cm cube on a 26\,$\times$\,18\,cm tray with a 1.2\,cm rim. We drew 41 layouts with a fixed random seed, placing the cube anywhere in a 16\,$\times$\,16\,cm zone and the tray 22--32\,cm from it, within 25$^{\circ}$ of straight left, and marked them on a sheet taped to a counter. One operator recorded each layout once with each device, in blocks of five layouts, alternating which device recorded a block first. In each take, the operator walks up to a floor mark while looking at the objects, brings the right hand in, pinches the cube between thumb and index finger, carries it onto the tray, lets go, and holds still for about 2\,s. The parts before the walk-up and after the task were trimmed off, leaving 16--29\,s per take.

For this experiment, EgoKit-PICO recorded both see-through cameras of the PICO 4 Ultra side by side (2560\,$\times$\,960 at 30\,fps, 76$^{\circ}$\,$\times$\,61$^{\circ}$ per eye), together with the head pose, the 26-joint hand tracking, and the intrinsics and extrinsics of both cameras. The Aria glasses recorded with HumanEgo's recommended settings (RGB at 1408\,$\times$\,1408 and 30\,fps), and each take went through MPS for SLAM and hand tracking. The 41 takes hold 14.6\,min of footage from the PICO and 14.1\,min from the Aria.

\subsection{Protocol}

The rules of the comparison, including its primary outcome, success on the shared layouts at 40 training takes, were fixed before the takes were recorded. The preprocessing settings, the training recipe, and the simulator are the same for both devices, except for a few values that depend on how a device reports the hand and the scene, such as the window in which the objects are located (1\,s for the stereo PICO and 2\,s for the single-camera Aria) and the gripper's pitch relative to the recorded hand frame. Each of these was set from that device's five pilot takes by a rule shared by both devices, never from how a policy performed. Policies were trained with the authors' recipe (400 epochs) on the first 10, 20, and 40 takes after take 000, which is held out; the 40-take condition was trained with three seeds. Every take was used, including those that missed a quality gate.

We assess the data before any learning and the policies after it. The \emph{separation error} is the distance between the cube and the tray as the pipeline locates them, minus their distance on the layout sheet. The \emph{quality gates} check each take, e.g., that the objects stay in the image, that the hand is tracked in at least 95\,\% of the frames between its entry and the release, and that the keypoints drawn into the training images sit on their object in at least 90\,\% of the frames. A \emph{replay} makes the simulated arm follow a take's own recorded hand path and grasp label in the scene rebuilt from that take. A policy \emph{succeeds} when the cube rests on the tray as its control loop ends, after at most 110 iterations. Each policy runs on the 41 recorded layouts of its device and on 100 shared layouts that neither device recorded, drawn from the same ranges with another seed; for these, the camera is placed at the median of the device's training-take camera poses relative to the layout. Because the policies of both devices run on the same shared layouts, we compare them pairwise: only the layouts where exactly one policy succeeds count (two-sided exact sign test), and a difference under about 15 points is reported as no difference detected.

\begin{table}[t]
\centering
\caption{HumanEgo on the cube-on-tray task, with demonstrations recorded on the same 41 layouts by EgoKit-PICO on PICO 4 Ultra and, as the reference, by Project Aria (Gen 1). Unless noted, policies were trained with seed 7.}
\label{tab:policy}
\footnotesize
\begin{tabularx}{\columnwidth}{@{}Xcc@{}}
\toprule
 & \textbf{PICO 4 Ultra} & \textbf{Aria Gen 1} \\
\midrule
\multicolumn{3}{@{}l}{\textit{Data, 41 takes per device}} \\
Footage (min) & 14.6 & 14.1 \\
Takes passing all quality gates & 33 & 39 \\
Separation error, mean (SD) (cm) & $+0.25$ (0.49) & $+0.08$ (0.38) \\
Pinch offset, along / across (cm)$^{1}$ & $-0.9$ / 1.6 & $+1.8$ / 2.6 \\
Takes whose replay places the cube & 29 & 33 \\
\midrule
\multicolumn{3}{@{}l}{\textit{Open-loop error on the held-out take (mm)}} \\
Policy, 10 / 20 / 40 takes & 25 / 26 / 28 & 34 / 30 / 34 \\
Hand stays still & 47 & 37 \\
\midrule
\multicolumn{3}{@{}l}{\textit{Cube set down, 100 unseen layouts}} \\
10 takes & 35 & 59 \\
20 takes & 70 & 66 \\
40 takes, seeds 7 / 8 / 9 & 67 / 74 / 60 & 73 / 78 / 0$^{2}$ \\
\midrule
\multicolumn{3}{@{}l}{\textit{Cube set down, 41 recorded layouts}} \\
10 / 20 / 40 takes & 16 / 29 / 33 & 17 / 31 / 36 \\
\bottomrule
\end{tabularx}
\\[2pt]
\begin{flushleft}\footnotesize
$^{1}$Median offset of the pinch point between the tracked thumb and index fingertips from the located cube when the grasp closes, along and across the camera's line of sight; negative is nearer the camera.\\
$^{2}$This policy declared the task finished at its first step in every episode.
\end{flushleft}
\end{table}

\begin{figure*}[t]
    \centering
    \includegraphics[width=\linewidth]{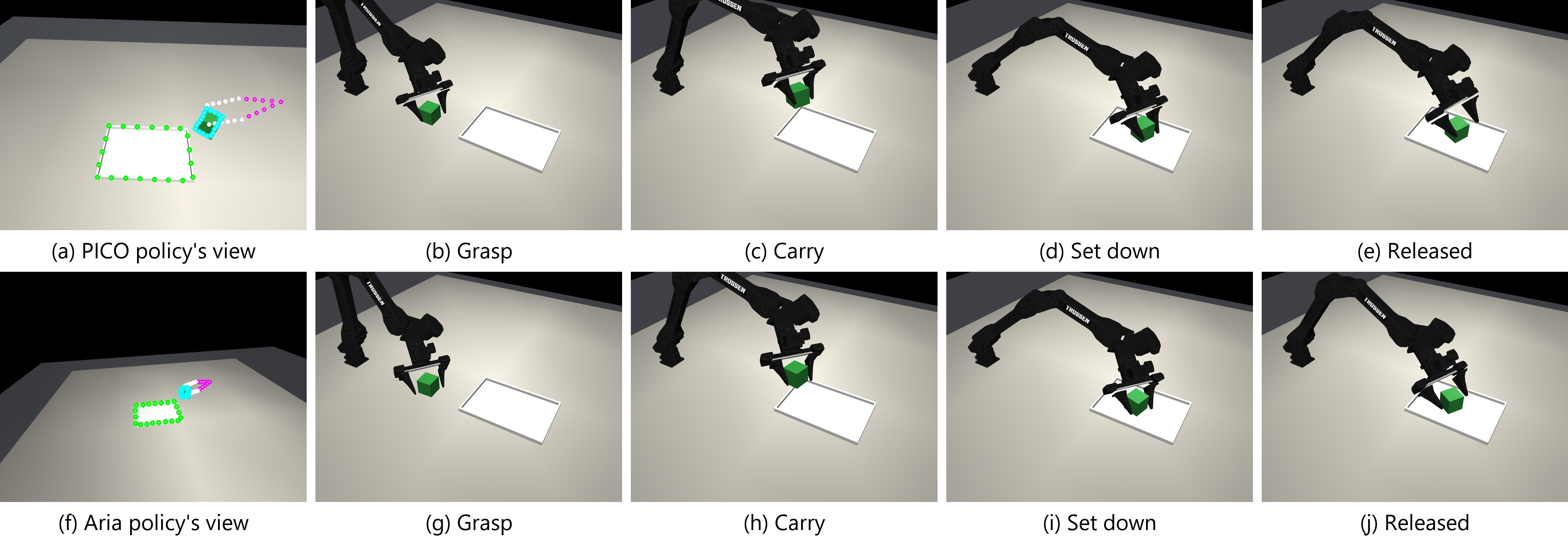}
    \caption{The 40-take policies of both devices (seed~7) in simulation, on one of the 100 shared layouts, which neither device recorded. (a, f) The image each policy is given at the moment of (c, h): the simulated scene from that device's camera, without the arm, with the virtual gripper and the keypoints drawn as in the training images. (b--e, g--j) An observer's view of the simulated arm following the policy: it grasps the cube, carries it, sets it down on the tray, and opens the gripper. Opening the gripper is not required for success, and the policies did not always do it: of the episodes in which they set the cube down (67 and 73), the PICO policy opened the gripper itself in 9 and the Aria policy in 24; in the others, the gripper still held the cube when the loop ended.}
    \label{fig:humanego_sim}
\end{figure*}

\subsection{Results}

Table~\ref{tab:policy} summarizes the results. Both devices' data locate the objects well: the mean separation error is 2.5\,mm for the PICO and 0.8\,mm for the Aria. Thirty-three PICO takes and 39 Aria takes pass every quality gate. In four of the eight PICO takes that miss a gate, the cube leaves the image for part of the take: the PICO's cameras cover a narrower field of view than the Aria's (76$^{\circ}$ against 90$^{\circ}$ horizontally in the processed images), so the operator has to keep the objects nearer the centre of the view. Two more PICO takes miss on hand tracking and two on the drawn keypoints; both Aria takes that miss do so on hand tracking. When the grasp closes, the pinch point between the tracked thumb and index fingertips lies nearer the located cube with the PICO than with the Aria, so the headset's on-device hand tracking agrees with the located objects at least as well as MPS does. Replaying the recorded hand paths puts the cube on the tray in 29 PICO and 33 Aria takes.

The policies learned the task from the data of both devices. On the held-out take, the policies trained on 10, 20, and 40 PICO takes predict the hand trajectory with a mean error of 25--28\,mm, against 47\,mm for a hand that stays still; the Aria policies 30--34\,mm against 37\,mm. In simulation, all policies trained on 20 or 40 takes but one set the cube down in 60 to 78 of the 100 unseen layouts; Fig.~\ref{fig:humanego_sim} shows the two 40-take policies on one of these layouts. With 10 takes, the Aria policy is ahead (59 against 35, one seed each). At 40 takes, the paired comparisons detect no difference between the devices: the PICO policy alone succeeds in 16 layouts and the Aria policy alone in 22 with seed 7 ($p=0.42$), and 12 against 16 with seed 8 ($p=0.57$). The exception is the Aria policy of seed 9, which declares the task finished at its first step in every episode and never moves the arm. The policies do not transfer between the devices: on the other device's held-out take, their error is 77--174\,mm, more than that of a hand that stays still.

In short, demonstrations recorded with EgoKit-PICO, with tracking from the headset itself rather than from a cloud service, trained HumanEgo policies that perform the task in simulation about as often as policies trained on recordings from the device the method was designed for. The validation has limits. It runs in simulation, which supplies the object masks, so it scores the motion rather than the perception of a deployed system. The simulated rig is ours, not the authors' robot, and the arm stands at the same place relative to the cube in every layout, so reach is not tested. The objects' orientations are assumed rather than measured, and all takes come from one operator in one room on one task.

\section{Discussion}

We aim to devise a unified low-cost toolkit so as to help scale up egocentric data collection. The barriers across different platforms bring several unexpected challenges. For example, smart glasses---Project Aria glasses in this case---do not provide mobile SDKs. As a result, we have to make a trade-off, namely calling the Aria application on Android through the accessibility feature on Android. In this way, there is no need to manually tap the record button in the Aria app to start the ego view recording and then open another app to record wrist views. In addition, because of the lack of a mobile SDK, we were unable to preview the ego view as in other EgoKit applications.

Ego view recording is easy on most devices, such as phones and headsets. It becomes challenging when recording the wrist view in a portable manner without relying on an additional PC or laptop. Moreover, OS differences are also troublesome. For example, iPadOS 18.7.1 supports USB cameras, while iOS 26.4.2 does not. It is worth mentioning that a jailbroken iPhone could support USB cameras~\cite{Workingw91:online}, but this has not been tested. Another Apple-family device, Apple Vision Pro, running visionOS 26.3, can support USB cameras and up to one USB hub, but it does not support two USB cameras simultaneously. It can switch the USB slot at intervals, but the frame rate of the two recorded wrist cameras is very low ($<$ 1\,fps). There might be another applicable scenario, namely recording a single wrist view to collect ego-view plus one-wrist-view data for training a single-arm model. Moreover, unlike Meta Quest, which exposes raw camera access via \textbf{PassthroughCameraAccess}, and PICO 4 Ultra, which provides raw camera access via \textbf{PXR\_CameraImage} based on the OpenXR extension \textbf{XR\_PICO\_camera\_image}, Apple Vision Pro requires \textbf{Enterprise APIs} for raw camera access. However, Enterprise APIs for visionOS are eligible for business use only. Per Apple, \textit{to be eligible to request the entitlement, your app needs to: (1) be for use in a business setting only; and (2) meet specific criteria associated with usage for each API.} The entitlement request is unavailable even for individuals in the Apple Developer Program. In our implementation, the ego view is obtained through the internal view recording function. To approximate raw camera access, the UI is hidden when wrist view recording starts. Besides, purchasing the Apple Vision Pro Developer Strap also requires a developer account. These restrictions render Apple Vision Pro less suitable for egocentric data collection. A similar work~\cite{park2024avp} uses a 3D-printed mount placed on the headset for ego view recording.

Headsets natively support head tracking and hand tracking, but smart glasses---Project Aria Gen 1 glasses---do not support real-time head tracking and hand tracking. The Machine Perception Services (MPS) cannot run locally. Phone-based applications can utilize open-source frameworks, such as MediaPipe, to detect hand landmarks in real time or after recording. The phone's 6-DoF pose can also be obtained via ARKit on iPhone or ARCore on Android phones.

As most devices have only one USB-C port, the power supply for the wrist cameras and the host device must be considered for long-duration recordings. The setup in this work uses an off-the-shelf USB-C hub, which does not support power input. To scale up egocentric data collection in a portable setting, this factor should be taken into consideration.


\section{Limitations and Future Work}

We use consumer-grade USB cameras and a wrist mount to record the wrist views. Since this mount is not designed for the data collection scenario, its structure is not compact. The mount and the camera collide with the desk when the hands are too close to it. Moreover, because of the mount position, the camera's view is partially blocked by the hand. A fish-eye camera with a customized compact mount would enable better wrist view recording.

Since the USB-C hub and the wrist cameras draw power from the host device---phone, iPad, or headset---power consumption must be considered, as it drains the battery faster. A customized USB-C hub that supports power input to the host device would enable long-duration recording. This is non-trivial when scaling up egocentric data collection in a factory setting.

EgoKit supports raw data collection, including the ego view and wrist views, as well as additional head and hand tracking data for XR headsets such as Quest 3 and PICO 4 Ultra. One issue inherent in the heterogeneous hardware and software is precise time synchronization, which deserves further investigation. One simple workaround for the current version is to let the three cameras simultaneously capture a millisecond-level timer displayed on a webpage; the views can then be aligned by processing the frames. Another issue is the frame rate difference, which can be resolved by sampling. For XR headsets, the head and hand tracking also need to be synchronized with the cameras.

Post-processing is always available for ego data collection; however, it can benefit greatly from early intervention, such as real-time feedback and real-time evaluation via on-device AI models. For example, during data collection, data quality factors such as lighting, occlusion, and hand tracking can be evaluated. If the data quality is low or hand tracking is off, the application can provide auditory feedback so that the collector can make adjustments. This can improve overall data quality and save time and resources compared with post-processing. Moreover, voice interaction is preferable to physical button presses, since the operator's hands are occupied. Voice-first interaction is feasible via simple keyword detection using on-device AI models.

Section~\ref{sec:policy} validated EgoKit-PICO data for policy learning in simulation, on one task recorded by one operator. Natural next steps are to deploy such policies on a real robot, to extend the validation to more tasks, operators, and EgoKit host devices, to fine-tune a VLA model or a world model on EgoKit data, and to validate EgoKit on a large-scale dataset collected with it.

\section{Conclusion}


We present EgoKit, a toolkit that exposes the same egocentric recording workflow across six heterogeneous host devices. Rather than designing new hardware, EgoKit ships per-platform applications that share a common recording interaction, a uniform log format, and on XR headsets synchronized head tracking and OpenXR-standard 26-joint hand tracking, paired with off-the-shelf consumer-grade accessories that extend any supported host with two USB cameras at a total cost of about \$151. We investigated the OS-level and SDK-level constraints encountered on each platform and reported the resulting video and tracking characteristics per device. To validate that EgoKit data can train robot policies, we replicated HumanEgo with demonstrations recorded by EgoKit-PICO: in simulation, the resulting policies placed a cube on a tray about as often as policies trained on recordings from Project Aria, the device HumanEgo was designed for.

Across the host devices we evaluated, XR headsets offered the most complete egocentric capture envelope, providing on-device head tracking and hand tracking alongside ego-view and wrist-view video in a self-contained form factor. Among the three tested headsets, PICO 4 Ultra was the least restricted in practice: Meta Quest 3 requires a Horizon OS version below 2.x and offers limited sensor access, and Apple Vision Pro reserves raw scene-camera access behind enterprise entitlements and does not support two USB cameras simultaneously. These observations are not intended as a ranking of the underlying hardware, but as guidance for teams selecting hardware for egocentric data collection under current constraints, which we expect to shift with future platform updates.

\bibliographystyle{IEEEtran}
\bibliography{main.bib}

\end{document}